\documentclass{article}

\usepackage[final]{nips_2018}
\usepackage{graphicx}
\usepackage{bm}
\usepackage{adjustbox}
\usepackage{rotating}
\usepackage{geometry}
\usepackage{enumitem,kantlipsum}

\usepackage[utf8]{inputenc} 
\usepackage[T1]{fontenc}    
\usepackage{hyperref}       
\usepackage{url}            
\usepackage{booktabs}       
\usepackage{amsfonts}       
\usepackage{nicefrac}       
\usepackage{microtype}      

\title{On the Importance of Strong Baselines \\ in Bayesian Deep Learning}

\author{
  Jishnu Mukhoti \\
  Department of Computer Science \\
  University of Oxford \\
  \texttt{jishnu.mukhoti@cs.ox.ac.uk} \\
  \And
  Pontus Stenetorp \\
  Department of Computer Science \\
  University College London \\
  \texttt{p.stenetorp@cs.ucl.ac.uk} \\
  \AND
  Yarin Gal \\
  Department of Computer Science \\
  University of Oxford \\
  \texttt{yarin@cs.ox.ac.uk} \\
}

\begin{document}

\maketitle

\begin{abstract}

Like all sub-fields of machine learning Bayesian Deep Learning is driven by empirical validation of its theoretical proposals.
Given the many aspects of an experiment it is always possible that minor or even major experimental flaws can slip by both authors and reviewers.
One of the most popular experiments used to evaluate approximate inference techniques is the regression experiment on UCI datasets.
However, in this experiment, models which have been trained to convergence have often been compared with baselines trained only for a fixed number of iterations.
We find that a well-established baseline, Monte Carlo dropout, when evaluated under the same experimental settings shows significant improvements.
In fact, the baseline outperforms or performs competitively with methods that claimed to be superior to the very same baseline method when they were introduced.
Hence, by exposing this flaw in experimental procedure, we highlight the importance of using identical experimental setups to evaluate, compare, and benchmark methods in Bayesian Deep Learning.
\end{abstract}

\section{Introduction}
Empiricism is at the very core of machine learning research, where we demand that new methods and approaches compare favorably to previously introduced work.
We expect this in terms of performance on either artificially generated data that highlights specific challenges and/or real-world data for specific tasks.
In this process, we implicitly rely on fellow scientists and reviewers to note discrepancies (intentional or not, to err is after all human) in the experimental setting -- such as for example, any kind of overfitting.\footnote{\url{http://hunch.net/?p=22}}
Recently, several studies have noted empirical shortcomings in the machine learning literature.
For example, \cite{henderson2017deep} observed that due to non-determinism, variance, and lack of significance metrics, it is difficult to judge whether claimed advances in reinforcement learning are empirically justified.
Also, \cite{melis2018on} established that several years of assumed progress in language modeling did not in fact improve upon a standard stacked LSTM model if the hyperparameters for all models were adequately tuned.
%

%
Bayesian Deep Learning applies the ideas of Bayesian inference to deep networks and is an active area of machine learning research.
Popular techniques for approximate inference in deep networks include variational inference (VI) \citep{graves2011practical}, probabilistic backpropagation (PBP) \citep{hernandez2015probabilistic} for Bayesian neural nets, dropout as an interpretation of approximate Bayesian inference \citep{Gal2016Dropout}, Deep Gaussian Processes (DGP) as a multi-layer generalization of Gaussian Processes \citep{bui2016deep}, Bayesian neural networks using Variational Matrix Gaussian (VMG) posteriors \citep{louizos2016structured},
and variants of Stochastic Gradient Hamiltonian Monte Carlo (SGHMC) methods \citep{springenberg2016bayesian}.
A well-defined experimental setup is necessary to compare and benchmark these methods; one of the most popular setups is a regression experiment on a number of curated UCI datasets \citep{hernandez2015probabilistic}.
The choice of this experiment is appealing for Bayesian neural nets because it provides predictive log-likelihood in addition to RMSE as an evaluation metric.
As such, predictive log-likelihood can be used to judge the quality of the uncertainty estimates produced by the model \citep{Gal2016Dropout}.
%

%
%
In this study, we observe a discrepancy in the setup used for the UCI regression experiment in some recent works, in that they compare their model to baselines obtained under a different experimental setting.
Concretely, this applies to VMG \citep{louizos2016structured}, HS-BNN \citep{ghosh2017model}, PBP-MV, \citep{sun2017learning} and SGHMC \citep{springenberg2016bayesian}.
In order to gauge the impact of this erroneous comparison, we reevaluate the regression experiments of the above-mentioned works compared to MC dropout \citep{Gal2016Dropout} in the same setting.
The experimental results indicate that the networks with dropout inference -- when trained under the same conditions -- outperform VMG, HS-BNN, and SGHMC; and they are a close second to PBP-MV.
These results suggest that several methods, when introduced, erroneously claimed state of the art performance in their publications.

\section{Regression Experiments}
We perform the non-linear regression experiments proposed in \cite{hernandez2015probabilistic}, which have been adopted to evaluate approximate inference techniques in numerous subsequent works: \cite{Gal2016Dropout}, \cite{bui2016deep}, \cite{louizos2016structured}, \cite{ghosh2017model}, \cite{sun2017learning}, \cite{springenberg2016bayesian} etc.
All the datasets from \cite{hernandez2015probabilistic} are used, apart from the \textit{YearPredictionMSD} dataset.
The \textit{YearPredictionMSD} dataset is very large with 515,345 instances, each of which has 90 dimensions.
Hence, tuning network hyperparameters over this dataset requires an inordinate amount of time.
Our network architecture has a single hidden layer with 50 hidden units for all datasets, except for the \textit{Protein Structure} dataset for which there are 100 hidden units.
There are two ways in which this experiment has been conducted in the past.
In the first, the networks are trained for a fixed number of iterations (specifically, 40 epochs), and the average training time of the networks are noted and compared.
This setting was used by \cite{hernandez2015probabilistic}, \cite{Gal2016Dropout}, and \cite{bui2016deep}.
In this work, we refer to this as the \textit{timed setting} of the experiment.
In the second variant of the experiment, the networks are trained to convergence with a higher number of training iterations and was used by \cite{louizos2016structured}, \cite{ghosh2017model}, \cite{sun2017learning}, \cite{springenberg2016bayesian}, and others.
For both variants, the test set RMSE and log-likelihood values are used as the evaluation metrics.
Given these two settings, it is a natural question to ask if models trained under the timed setting and those trained to convergence are comparable.
One might argue that networks trained for a fixed number of iterations might not have converged and will thus perform poorly compared to those which have been trained to convergence.
To test this hypothesis, we use networks with MC dropout \citep{Gal2016Dropout} -- one of the de facto standard baselines for approximate inference.
%

%
%
%
%

We compare with the following works:
VMG \citep{louizos2016structured}, where models trained to convergence have been benchmarked against MC dropout baselines from the timed setting;
Bayesian networks with horseshoe priors (HS-BNN) \citep{ghosh2017model} and probabilistic backpropagation with the Matrix Variate Gaussian (MVG) distribution (PBP-MV) \citep{sun2017learning}, where the models were only benchmarked against VMG;
Stochastic Gradient Hamiltonian Monte Carlo methods (SGHMC) \citep{springenberg2016bayesian}, where the results have been compared with Probabilistic Backpropagation (PBP) \citep{hernandez2015probabilistic} and Variational Inference (VI) \citep{graves2011practical} baselines obtained under the timed setting.

There are two hyperparameters: i) the model precision parameter $\tau$ to evaluate the log-likelihood and ii) the dropout rate $d$.
We perform the following two variations of the regression experiment:
\begin{enumerate}[leftmargin=*]
  \item{
  	\textbf{Convergence}: The networks are trained to convergence for 4,000 epochs and the hyperparameter values are obtained by Bayesian Optimization (BO) (as described in \cite{Gal2016Dropout}).
  }
  \item{
  	\textbf{Hyperparameter tuning}: Just as in the previous variant the networks are trained for 4,000 epochs, but we also obtain optimal hyperparameter values by performing grid search over a range of $(\tau, d)$ pairs and choose the best pair based on performance over a validation set.
    The validation set is created by randomly choosing 20\% of the data points in the training set.
  }
\end{enumerate}
The RMSE and log likelihood values obtained from the above experiments are compared in Table~\ref{table1} and \ref{table2} respectively.
It should be noted that the RMSE and log likelihood values of VMG, HS-BNN, PBP-MV, and SGHMC have been taken from their respective papers.
The experimental results indicate that the \textit{Convergence} and \textit{Hyperparameter tuning} baseline experiments show a substantial improvement in performance compared to the results in the \textit{timed setting}.
We also observe that the baseline outperform the other methods on the \textit{Concrete Strength}, \textit{Naval Propulsion Plants}, \textit{Wine Quality Red}, and \textit{Yacht Hydrodynamics} datasets in terms of RMSE.
With respect to log likelihood, the baseline performs best on the \textit{Boston Housing}, \textit{Concrete Strength}, and \textit{Wine Quality Red} datasets.
Finally, we also observe that on the other datasets the baseline is competitive -- second only to PBP-MV \citep{sun2017learning}.
%

%
%
%
%

%
\begin{table}
  \caption{\textbf{Average RMSE test performance.} The RMSE values along with corresponding standard errors are presented.}
  \label{table1}
  \centering
  \resizebox{\textwidth}{!}{
  \begin{tabular}{c|c|c|c|c|c|c}
    \toprule
    \textbf{Dataset} & \begin{tabular}{@{}c@{}}\textbf{Dropout} \\ \textbf{(Timed Setting)} \end{tabular} & \begin{tabular}{@{}c@{}}\textbf{Dropout} \\ \textbf{(Convergence)} \end{tabular} & \begin{tabular}{@{}c@{}}\textbf{Dropout} \\ \textbf{(Hyperparameter} \\ \textbf{tuning)} \end{tabular} & \begin{tabular}{@{}c@{}}\textbf{VMG} \end{tabular} & \begin{tabular}{@{}c@{}}\textbf{HS-BNN} \end{tabular} & \begin{tabular}{@{}c@{}}\textbf{PBP-MV} \end{tabular} \\
    \hline
    Boston Housing & $2.97 \pm 0.19$ & $2.83 \pm 0.17$ & $2.90 \pm 0.18$ & $\bm{2.70 \pm 0.13}$ & $3.32 \pm 0.66$ & $3.11 \pm 0.15$ \\
    Concrete Strength & $5.23 \pm 0.12$ & $4.93 \pm 0.14$ & $\bm{4.82 \pm 0.16}$ & $4.89 \pm 0.12$ & $5.66 \pm 0.41$ & $5.08 \pm 0.14$ \\
    Energy Efficiency & $1.66 \pm 0.04$ & $1.08 \pm 0.03$ & $0.54 \pm 0.06$ & $0.54 \pm 0.02$ & $1.99 \pm 0.34$ & $\bm{0.45 \pm 0.01}$ \\
    Kin8nm & $0.10 \pm 0.00$ & $0.09 \pm 0.00$ & $0.08 \pm 0.00$ & $0.08 \pm 0.00$ & $0.08 \pm 0.00$ & $\bm{0.07 \pm 0.00}$ \\
    Naval Propulsion & $0.01 \pm 0.00$ & $\bm{0.00 \pm 0.00}$ & $\bm{0.00 \pm 0.00}$ & $\bm{0.00 \pm 0.00}$ & $\bm{0.00 \pm 0.00}$ & $\bm{0.00 \pm 0.00}$ \\
    Power Plant & $4.02 \pm 0.04$ & $4.00 \pm 0.04$ & $4.01 \pm 0.04$ & $4.04 \pm 0.04$ & $4.03 \pm 0.15$ & $\bm{3.91 \pm 0.04}$ \\
    Protein Structure & $4.36 \pm 0.01$ & $4.27 \pm 0.01$ & $4.27 \pm 0.02$ & $4.13 \pm 0.02$ & $4.39 \pm 0.04$ & $\bm{3.94 \pm 0.02}$ \\
    Wine Quality Red & $0.62 \pm 0.01$ & $\bm{0.61 \pm 0.01}$ & $0.62 \pm 0.01$ & $0.63 \pm 0.01$ & $0.63 \pm 0.04$ & $0.64 \pm 0.01$ \\
    Yacht Hydrodynamics & $1.11 \pm 0.09$ & $0.70 \pm 0.05$ & $\bm{0.67 \pm 0.05}$ & $0.71 \pm 0.05$ & $1.58 \pm 0.23$ & $0.81 \pm 0.06$ \\
    \bottomrule
  \end{tabular}}
\end{table}

\begin{table}  
  \caption{\textbf{Average log likelihood test performance.} The log likelihood values along with corresponding standard errors are presented.}
  \label{table2}
  \centering
  \resizebox{\textwidth}{!}{
  \begin{tabular}{c|c|c|c|c|c|c|c|c}
    \toprule
    \textbf{Dataset} & \begin{tabular}{@{}c@{}c@{}}\textbf{Dropout} \\ \textbf{(Timed} \\ \textbf{Setting)} \end{tabular} & \begin{tabular}{@{}c@{}}\textbf{Dropout} \\ \textbf{(Convergence)} \end{tabular} & \begin{tabular}{@{}c@{}}\textbf{Dropout} \\ \textbf{(Hyperparameter} \\ \textbf{tuning)} \end{tabular} & \begin{tabular}{@{}c@{}}\textbf{VMG} \end{tabular} & \begin{tabular}{@{}c@{}}\textbf{HS-BNN} \end{tabular} & \begin{tabular}{@{}c@{}}\textbf{PBP-MV} \end{tabular} & \begin{tabular}{@{}c@{}c@{}}\textbf{SGHMC} \\  \textbf{(Tuned per} \\ \textbf{dataset)} \end{tabular} & \begin{tabular}{@{}c@{}c@{}}\textbf{SGHMC} \\  \textbf{(Scale} \\ \textbf{Adapted)} \end{tabular} \\
    \hline
    Boston Housing & $-2.46 \pm 0.06$ & $\bm{-2.40 \pm 0.04}$ & $\bm{-2.40 \pm 0.04}$ & $-2.46 \pm 0.09$ & $-2.54 \pm 0.15$ & $-2.54 \pm 0.08$ & $-2.49 \pm 0.15$ & $-2.54 \pm 0.04$ \\
    Concrete Strength & $-3.04 \pm 0.02$ & $-2.97 \pm 0.02$ & $\bm{-2.93 \pm 0.02}$ & $-3.01 \pm 0.03$ & $-3.09 \pm 0.06$ & $-3.04 \pm 0.03$ & $-4.17 \pm 0.72$ & $-3.38 \pm 0.24$ \\
    Energy Efficiency & $-1.99 \pm 0.02$ & $-1.72 \pm 0.01$ & $-1.21 \pm 0.01$ & $-1.06 \pm 0.03$ & $-2.66 \pm 0.13$ & $\bm{-1.01 \pm 0.01}$ & $--$ & $--$ \\
    Kin8nm & $0.95 \pm 0.01$ & $0.97 \pm 0.00$ & $1.14 \pm 0.01$ & $1.10 \pm 0.01$ & $1.12 \pm 0.03$ & $\bm{1.28 \pm 0.01}$ & $--$ & $--$ \\
    Naval Propulsion & $3.80 \pm 0.01$ & $3.91 \pm 0.01$ & $4.45 \pm 0.00$ & $2.46 \pm 0.00$ & $\bm{5.52 \pm 0.10}$ & $4.85 \pm 0.06$ & $--$ & $--$ \\
    Power Plant & $-2.80 \pm 0.01$ & $-2.79 \pm 0.01$ & $-2.80 \pm 0.01$ & $-2.82 \pm 0.01$ & $-2.81 \pm 0.03$ & $\bm{-2.78 \pm 0.01}$ & $--$ & $--$ \\
    Protein Structure & $-2.89 \pm 0.00$ & $-2.87 \pm 0.00$ & $-2.87 \pm 0.00$ & $-2.84 \pm 0.00$ & $-2.89 \pm 0.00$ & $\bm{-2.77 \pm 0.01}$ & $--$ & $--$ \\
    Wine Quality Red & $-0.93 \pm 0.01$ & $\bm{-0.92 \pm 0.01}$ & $-0.93 \pm 0.01$ & $-0.95 \pm 0.01$ & $-0.95 \pm 0.05$ & $-0.97 \pm 0.01$ & $-1.29 \pm 0.28$ & $-1.04 \pm 0.17$ \\
    Yacht Hydrodynamics & $-1.55 \pm 0.03$ & $-1.38 \pm 0.01$ & $-1.25 \pm 0.01$ & $-1.30 \pm 0.02$ & $-2.33 \pm 0.01$ & $-1.64 \pm 0.02$ & $-1.75 \pm 0.19$ & $\bm{-1.10 \pm 0.08}$ \\
    \bottomrule
  \end{tabular}}
\end{table}

\section{Conclusion}
The RMSE and log-likelihood values obtained from the \textit{Convergence} and \textit{Hyperparameter tuning} experiments (as given in Tables~\ref{table1} and \ref{table2}) provide substantially better results for MC dropout.
We conclude that previous comparisons with baselines obtained from the timed setting are unrepresentative, as the models in one setting had not reached convergence.
In summary, when benchmarking a method, its performance should always be evaluated using an experimental setup identical to the one used to evaluate its peers.
%

%
%
%
The source code for our experiments can be found at: \url{https://github.com/yaringal/DropoutUncertaintyExps}


\bibliography{refs}

\begin{thebibliography}{}

\bibitem[Bui et~al., 2016]{bui2016deep}
Bui, T., Hern{\'a}ndez-Lobato, D., Hernandez-Lobato, J., Li, Y., and Turner, R.
  (2016).
\newblock Deep gaussian processes for regression using approximate expectation
  propagation.
\newblock In {\em International Conference on Machine Learning}, pages
  1472--1481.

\bibitem[Gal and Ghahramani, 2016]{Gal2016Dropout}
Gal, Y. and Ghahramani, Z. (2016).
\newblock Dropout as a {B}ayesian approximation: Representing model uncertainty
  in deep learning.
\newblock In {\em Proceedings of the 33rd International Conference on Machine
  Learning (ICML-16)}.

\bibitem[Ghosh and Doshi-Velez, 2017]{ghosh2017model}
Ghosh, S. and Doshi-Velez, F. (2017).
\newblock Model selection in bayesian neural networks via horseshoe priors.
\newblock {\em arXiv preprint arXiv:1705.10388}.

\bibitem[Graves, 2011]{graves2011practical}
Graves, A. (2011).
\newblock Practical variational inference for neural networks.
\newblock In {\em Advances in neural information processing systems}, pages
  2348--2356.

\bibitem[Henderson et~al., 2017]{henderson2017deep}
Henderson, P., Islam, R., Bachman, P., Pineau, J., Precup, D., and Meger, D.
  (2017).
\newblock Deep reinforcement learning that matters.
\newblock {\em arXiv preprint arXiv:1709.06560}.

\bibitem[Hern{\'a}ndez-Lobato and Adams, 2015]{hernandez2015probabilistic}
Hern{\'a}ndez-Lobato, J.~M. and Adams, R. (2015).
\newblock Probabilistic backpropagation for scalable learning of bayesian
  neural networks.
\newblock In {\em International Conference on Machine Learning}, pages
  1861--1869.

\bibitem[Louizos and Welling, 2016]{louizos2016structured}
Louizos, C. and Welling, M. (2016).
\newblock Structured and efficient variational deep learning with matrix
  gaussian posteriors.
\newblock In {\em International Conference on Machine Learning}, pages
  1708--1716.

\bibitem[Melis et~al., 2018]{melis2018on}
Melis, G., Dyer, C., and Blunsom, P. (2018).
\newblock On the state of the art of evaluation in neural language models.
\newblock In {\em International Conference on Learning Representations}.

\bibitem[Springenberg et~al., 2016]{springenberg2016bayesian}
Springenberg, J.~T., Klein, A., Falkner, S., and Hutter, F. (2016).
\newblock Bayesian optimization with robust bayesian neural networks.
\newblock In {\em Advances in Neural Information Processing Systems}, pages
  4134--4142.

\bibitem[Sun et~al., 2017]{sun2017learning}
Sun, S., Chen, C., and Carin, L. (2017).
\newblock Learning structured weight uncertainty in bayesian neural networks.
\newblock In {\em Artificial Intelligence and Statistics}, pages 1283--1292.

\end{thebibliography}
\bibliographystyle{apalike}

\end{document}